\documentclass{article}

\usepackage[final]{neurips_2019}

\usepackage[utf8]{inputenc} 
\usepackage[T1]{fontenc}    
\usepackage{hyperref}       
\usepackage{url}            
\usepackage{booktabs}       
\usepackage{amsfonts}       
\usepackage{nicefrac}       
\usepackage{microtype}      

\usepackage{todonotes}
\usepackage{graphicx}
\usepackage{comment}
\usepackage{float}
\usepackage{bm}
\usepackage{hyperref}

\usepackage{amsmath}
\usepackage{xcolor}
\usepackage{natbib}
\usepackage{indentfirst}
\usepackage{wrapfig}
\usepackage{xcolor}
\usepackage{caption}
\usepackage{subcaption}

\newcommand{\Normal}{\textsc{Normal }}

\newcommand{\NotTrained}{\textsc{Untrained }}

\newcommand{\ShuffledPixelsNoSpace}{\textsc{Shuffled Pixels}}

\title{Neural Networks Trained on Natural Scenes\\Exhibit Gestalt Closure}

\author{Been Kim, Emily Reif, Martin Wattenberg, Samy Bengio, Michael C.~Mozer\\
\emph{Google Research}\\
\emph{1600 Amphitheater Parkway}\\
\emph{Mountain View, CA 94043}}

\begin{document}

\maketitle


%

%


\vskip 0.3in

\begin{abstract}

The Gestalt laws of perceptual organization, which describe how visual elements in an image are grouped  and interpreted, have traditionally been thought of as innate despite their ecological validity. We use deep-learning methods to investigate whether natural scene statistics might be sufficient to derive the Gestalt laws. We examine the law of \emph{closure}, which asserts that human visual perception tends to ``close the gap''  by assembling elements that can jointly  be interpreted as a  complete figure or object. We demonstrate that  a state-of-the-art convolutional neural network, trained to classify natural images, exhibits closure  on synthetic displays of edge fragments, as assessed by similarity of internal representations. This  finding provides support for the hypothesis that the  human perceptual system is even more elegant than  the Gestaltists imagined: a single law---adaptation to  the statistical structure of the  environment---might suffice as fundamental.
\end{abstract}


Psychology has long aimed to discover fundamental laws of behavior that place 
the field on the same footing as `hard' sciences like physics and chemistry~\citep{schultz2015history}.
Perhaps the most visible and overarching set of such laws, developed
in the early twentieth century to explain perceptual and attentional 
phenomena, are the \emph{Gestalt principles}~\citep{wertheimer1923laws}. 
These principles have had a tremendous impact on modern 
psychology~\citep{kimchi1992primacy, wagemans2012century1, wagemans2012century2, schultz2015history}.
Although Gestalt psychology has faced some criticism over a lack of rigor~\citep{wagemans2012century1, westheimer1999gestalt, schultz2015history}, investigators have successfully operationalized 
its concepts~\citep{ren2003learning}, and it has influenced work in medicine~\citep{bender1938visual}, computer vision~\citep{desolneux2007gestalt}, therapy~\citep{zinker1977creative}, and 
design~\citep{behrens1998art}.

\begin{figure}[tb]
    \centering
    \includegraphics[height=1in]{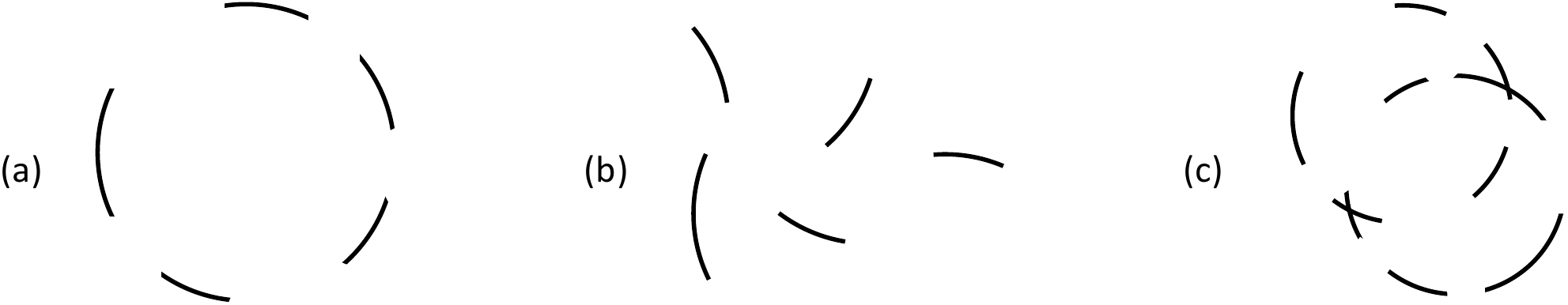}
    \caption{(a) A circle formed from fragments via closure; (b) the same fragments but rearranged to prevent closure;
    (c) fragments from two circles which can be segmented using closure }
    \label{fig:new_examples}
\end{figure}

The Gestalt principles describe how visual elements are grouped and interpreted.
For example, the Gestalt principle of \emph{closure} asserts that human visual 
perception tends to ``close the gap'' by grouping elements that can jointly 
be interpreted as a complete figure or object.
The principle thus provides a basis for predicting how viewers will parse and understand display fragments such as those in Figures~\ref{fig:new_examples}a, b. The linking of fragments such as those in Figure~\ref{fig:new_examples}a occurs
early in perceptual processing, hampering access to the constituent fragments  but facilitating rapid recognition
of the completed shape \citep{RensinkEnns1998}.

The Gestalt principles can support object perception by  
grouping together strongly interrelated features---features likely to belong to the same 
object, allowing features of that object to be processed apart from the features of other objects
(e.g., Figure~\ref{fig:new_examples}c).
Consistent with this role of grouping, the Gestalt 
principles have long been considered to have ecological validity in the natural world 
\citep{Brunswik1953}. That is, natural image statistics have been shown to justify 
many of the Gestalt principles, including good continuation, proximity, and similarity
\citep{ElderGoldberg2002,Geisler2001,Kruger1998,Sigman2001}.

The Gestaltist tradition considered the principles to be innate and immutable~\citep{kimchi1992primacy}.
Although the role of learning was acknowledged, the atomic Gestalt principles were considered primary \citep{Todorovic2008}. 
Even the aforementioned research examining natural image statistics has presumed that the principles
either evolved to support ordinary perception or fortuitiously happen to have utility for perception.

%
%

\begin{figure}[tb]
    \centering
    \includegraphics[height=1in]{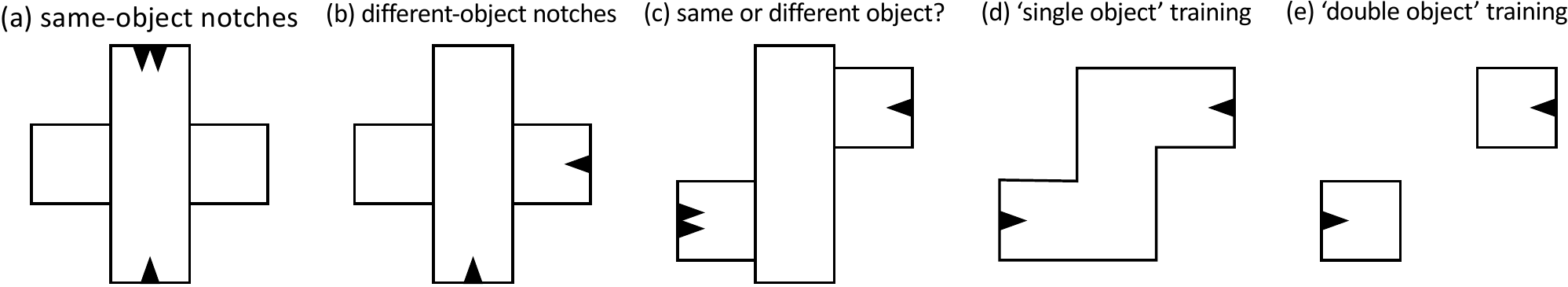}
    \caption{Examples of stimuli used by \citet{Zemel2002}}
    \label{fig:zemel}
\end{figure}

However, ample evidence supports the notion that perceptual grouping can be modulated by experience.  
For example, figure-ground segregation is affected by object familiarity: a silhouette is more 
likely to be assigned as the figure if it suggests a common object
\citep{PetersonGibson1994,Peterson2019}.
And perceptual grouping can be altered with only a small amount of experience in a novel 
stimulus environment \citep{Zemel2002}. In Zemel et al.'s studies, participants
were asked to report whether two features in a display matched. Consistent with 
\citet{Duncan1984}, participants are faster to respond
when the two features---notches on the ends of rectangles---belong to the 
same object (Figure~\ref{fig:zemel}a) relative to when they belong to different objects 
(Figure~\ref{fig:zemel}b).  Although participants treat Figures~\ref{fig:zemel}a,b as one rectangle occluding another,
the two small squares in Figure~\ref{fig:zemel}c are treated as distinct objects. However, following brief training on
stimuli such as the zig-zag shape in Figure~\ref{fig:zemel}d, the two small squares are treated as parts of the same object,
relative to a control condition in which the training consisted of fragments as in Figure~\ref{fig:zemel}e.

If perceptual grouping can be modulated by experience, perhaps the Gestalt principles are not innate and
immutable but rather are developmentally acquired as a consequence of interacting
with the natural world.
Ordinary perceptual experience might suffice to allow a learner to discover the Gestalt principles,
given that the statistical structure of the environment is consistent with the Gestalt principles
\citep{ElderGoldberg2002,Geisler2001,Kruger1998,Sigman2001}.
In the present work, we use deep-learning methods to investigate this hypothesis.

\begin{figure}[tb]
    \centering
    \includegraphics[width=5.5in]{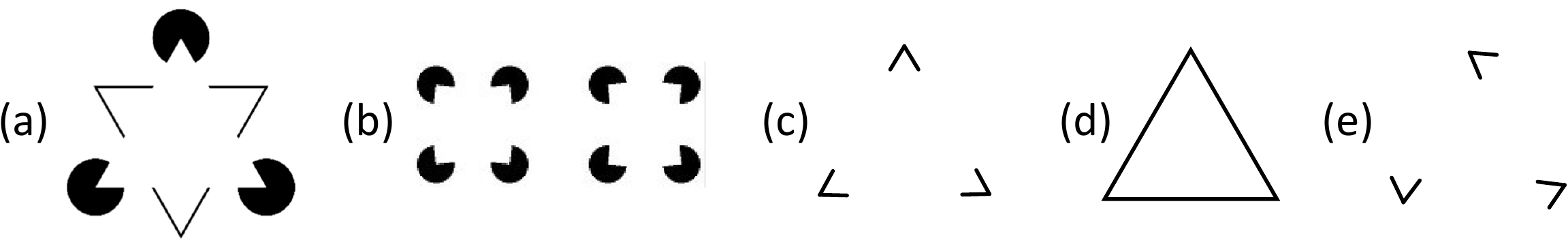}
    \caption{(a) The Kanisza triangle formed from illusory contours; 
    (b) fat and thin squares used as stimuli by \citet{Baker2018};
    (c) \emph{aligned} corners---the minimal visual cues required to induce closure and an illusory triangle;
    (d) a \emph{complete} triangle;
    and (e) \emph{disorderd} corners, which should be insufficient to induce closure or an illusory triangle.}
    \label{fig:kanizsa}
\end{figure}

We focus on closure (Figure~\ref{fig:new_examples}a,c). Closure is a particularly 
compelling illustration of the Gestalt perspective because fragments are assembled into a meaningful 
configuration and perceived as a unified whole \citep{wertheimer1923laws}. In this process, missing fragments 
can be filled in, resulting in illusory contours, the classic example of which is the Kanizsa 
triangle (Figure~\ref{fig:kanizsa}a). Traditional cognitive models have been designed to explain illusory contours
\citep[e.g.,][]{Grossberg2014,Kalar2010}. These models adopt the assumption of innateness in
that they are built on specialized mechanisms designed to perform some form of filling in.
We examine whether a deep  neural net trained on natural images exhibits closure effects naturally and as a
consequence of its exposure to its environment.

Closure has been examined experimentally in humans via measures that include response 
latency \citep{KIMCHI201634}, discrimination ability~\citep{ringach1996spatial}, and EEG response
\citep{sanguinetti2016increased}.
Measuring closure effects via behavior in feedforward neural 
networks is not entirely straightforward. These networks
have no temporal dynamics of response formation, so latency-based
response measures will be uninformative. And for discrimination
tasks, they require task-specific training, which complicates the use
of existing, pretrained models.

\citet{Baker2018} explored whether neural nets
`perceive' illusory contours using an indirect technique. They studied displays consisting of fragments
that could be completed as either fat or thin rectangles (Figure~\ref{fig:kanizsa}b, left and right images, respectively).
Using AlexNet \citep{Krizhevsky2012}, a pretrained network for image classification, they decapitated the output layer which represents every object class
and replaced it with a single unit that discriminates fat from thin rectangles. The weights from the penultimate layer to the output unit
were trained on complete (non-illusory) fat and thin rectangles presented in varying sizes, aspect ratios, and positions in the image.
This additional training extracts information available from the original model for fat versus thin classification.
Following training, the network could readily discriminate fat and thin rectangles, whether real or illusory.
\citet{Baker2018} then borrowed a method from the human behavioral literature, \emph{classification images} 
\citep{gold2000}, to infer features in the image that drive responses.  Essentially, the method adds pixelwise luminance noise to images 
and then uses an averaging technique to identify the pixels that reliably modulate the probability of a given response.
In humans, this method infers the illusory contours of the rectangles suggested by the stimuli in Figure~\ref{fig:kanizsa}b. 
In contrast, \citet{Baker2018} found no evidence that that pixels along illusory contours influence network classification decisions.

Although the classification-image paradigm finds a measurable difference between networks and humans, Baker et al.'s claim that
`deep convolutional networks do not perceive illusory contours' (the title of their article) is too strong.
Baker et al.\ treat networks as black boxes, which conflicts with their stated goal of probing
the nature of internal representations. One advantage of network experiments
over human experiments is that representations can be observed directly.  With humans, the classification-image paradigm is a necessary
and clever means of reverse-engineering representations; however, the paradigm recovers only \emph{image} representations not 
\emph{internal} representations. Illusory contours and filling-in processes should be manifested in internal representations.
This manifestation is likely not visual in nature: \citet{RensinkEnns1998} found no evidence that the inferred contours are superimposed on
a low-level visual representations; instead, they describe a type of functional or conceptual filling-in that influences 
high-level representations constructed from the low-level representation. This claim is consistent with monkey neurophysiology indicating 
that the first cortical stage (area 17) is not responsive to illusory contours whereas later stages (including area 18) 
are \citep{von1984illusory}.\footnote{Regardless of the issue of where and how illusory contours are manifested in the visual stream,
Baker et al.\ have identified a behavioral discrepancy between feedforward neural networks and humans which must be explained.
We return to this issue in the discussion.}





In our work, we examine the internal representations of a neural network trained on natural scenes and then tested on a minimalist 
instantiation of the Kanizsa triangle, which consists of three corner fragments
with the edges between them removed
(Figure~\ref{fig:kanizsa}c), similar to stimuli used in classic human studies
\citep[e.g.,][]{elder1993effect, KIMCHI201634, kimchi1992primacy}.
We investigate whether these
\emph{aligned} fragments yield a representation more similar to that of a  \emph{complete} triangle 
(Figure~\ref{fig:kanizsa}d) than do \emph{disordered} fragments (Figure~\ref{fig:kanizsa}e). 
Focusing on similarity of internal representations allows us to evaluate 
the Gestalt perception of shape and the functional filling in of the 
missing contours.
The aligned fragments should induce stronger illusory contours than do the disordered fragments.

\begin{figure}[tb!]
    \centering
    (a)
    \begin{subfigure}{.28\textwidth}
        \centering
        \includegraphics[width=.9\linewidth]{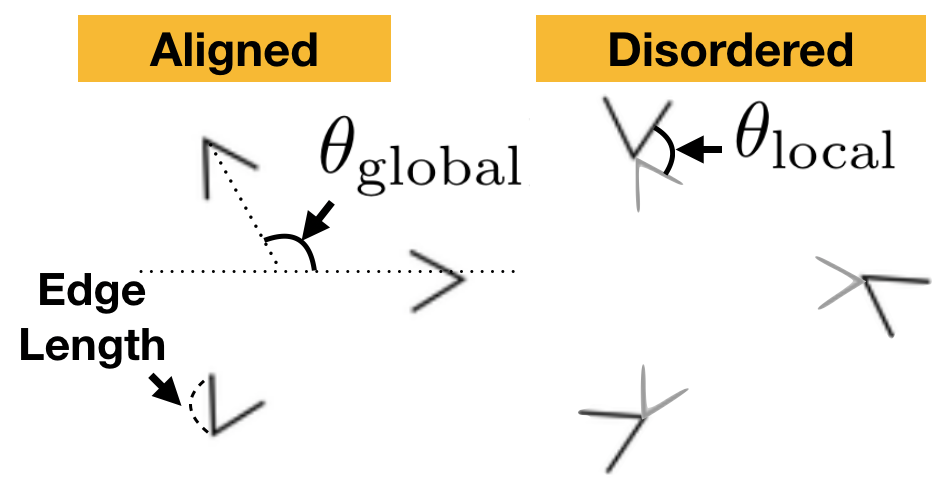}
    \end{subfigure}
    ~~~~~(b)
    \begin{subfigure}{.55\textwidth}
        \centering
        \includegraphics[width=.95\linewidth]{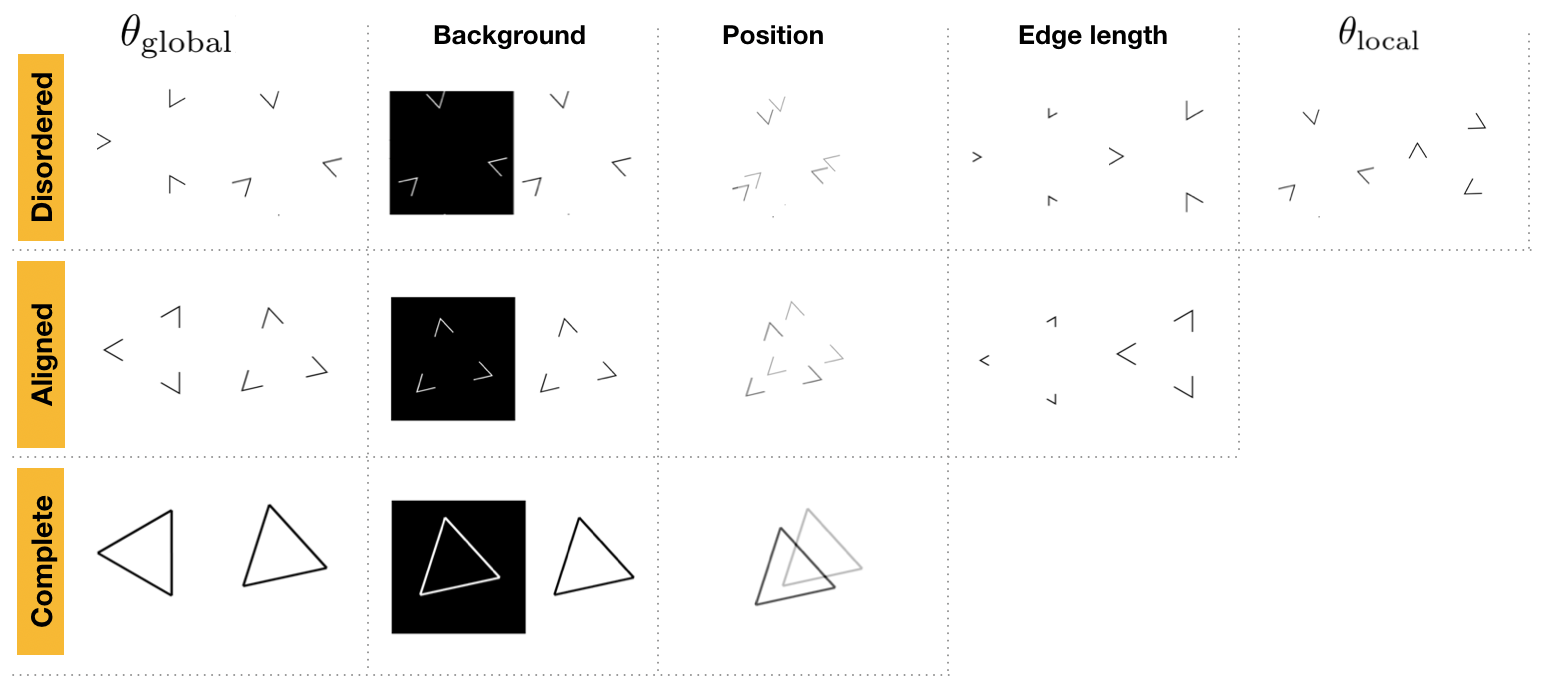}
    \end{subfigure}\\
   (c) \begin{subfigure}{.45\textwidth}
                \includegraphics[width=.9\linewidth]{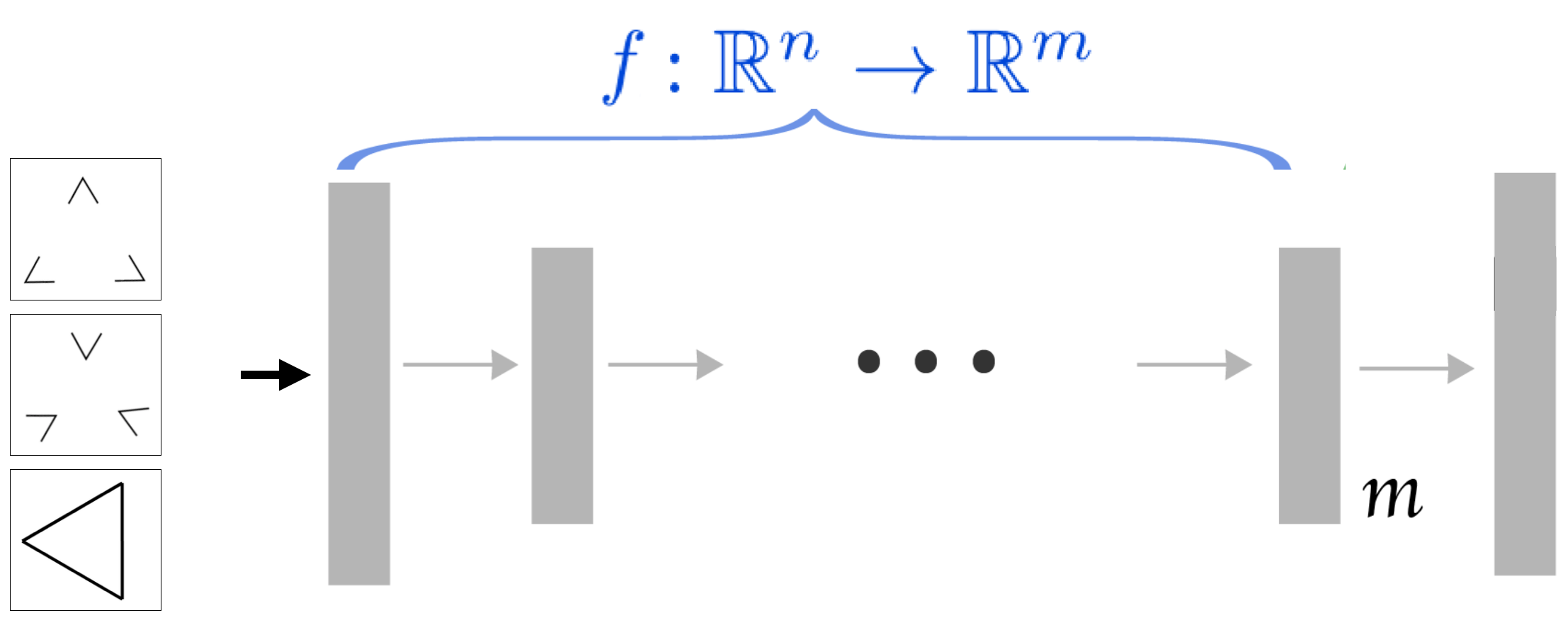}
                 \end{subfigure}
(d) \begin{subfigure}{.47\textwidth}
        \includegraphics[width=.9\linewidth]{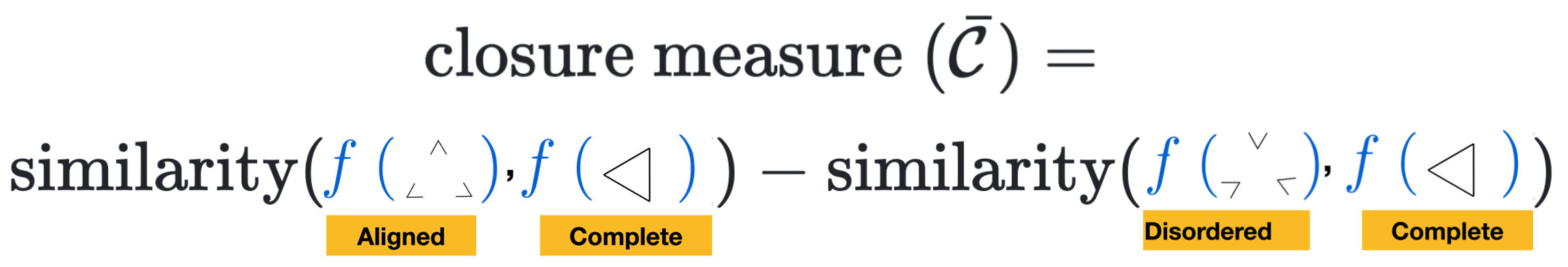}
            \end{subfigure}
    \caption{Outline of the stimuli and methodology to test closure in pretrained neural networks. (a) The tested
    shapes varied in global orientation and the disordered images also varied in local orientation of their 
    elements. (b) Examples of stimulus variation for the three experimental conditions (depicted in the rows), and for 
    five properties (depicted in the columns). (c) Images are fed into a deep ConvNet trained
    to classify images, where there is one output neuron per class.
    In most of our simulations, the penultimate layer, with $m$ units is used as a deep
    embedding of the input image. (d) Computing a closure measure, $\bar{\mathcal{C}}$, where a larger
    value indicates that the representation of the complete triangle is more similar to the
    representation of the aligned fragments than to the representation of the disordered fragments.
    Note that $\bar{\mathcal{C}}$ is an expectation over many image triples, not depicted in the equation.
    \label{fig:shape_properties}
    }
\end{figure}

Our stimuli are pixel images of complete, aligned, and disordered shapes, varying in a number of
irrelevant dimensions to ensure robustness of effects we observe (Figure~\ref{fig:shape_properties}a,b). 
The stimuli are fed into a pretrained deep convolutional neural net (hereafter,
\emph{ConvNet}) that maps the image to an
$m$-dimensional  internal representation, which we refer to as the \emph{embedding}
(Figure~\ref{fig:shape_properties}c).  We estimate the expected relative similarity of aligned and 
disordered fragments to the complete image using a closure measure, $\bar{\mathcal{C}} \in [-1,+1]$,
where a larger value indicates that the representation of the complete triangle is more like the
representation of the aligned fragments than the representation of the disordered fragments
(Figure~\ref{fig:shape_properties}d). 


\section*{Results} 



\subsection*{Sanity check}
We conduct a proof-of-concept experiment to show that we can distinguish models
that produce closure from those that do not.   To ensure that the models have these
properties, we train simple ConvNets from scratch solely on our set of complete, aligned, and 
disordered images. The networks are trained to perform one of two binary 
classification tasks: \emph{closure discrimination (CD)}, which produces output 1 for complete and aligned images and output 0 for disordered images, and 
\emph{background discrimination (BD)}, which produces output 1 for black backgrounds and
0 for white backgrounds. The CD net will necessarily treat complete and aligned as
more similar than complete and disordered, and should therefore produce a positive
$\bar{\mathcal{C}}$ score. In contrast, the BD net needs to extract color not 
shape information from images, and if it ignores shape altogether, it will 
yield a $\bar{\mathcal{C}}$ score of 0. Our aim in this contrast is to present the
pattern of results obtained in these two conditions as signatures of closure (for CD)
and lack of closure (for BD). 

\begin{figure}[tb]
\centering
    \includegraphics[width=.4\linewidth]{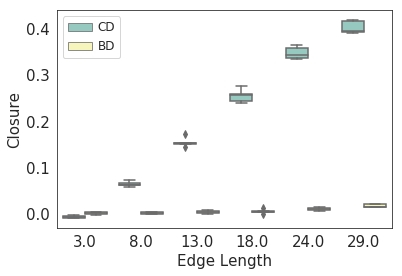}
    \caption{Sanity check experiment. CD networks, trained to discriminate complete and aligned from disordered images, show increasing closure with edge length.
    BD networks, trained to discriminate background color, show no closure.
    The shading around each line indicates a confidence interval obtained by
    running replications of each network. Only final layer is plotted.
        }
    \label{fig:sanity}
\end{figure}

Figure~\ref{fig:sanity} presents the closure measure ($\bar{\mathcal{C}}$) for
the CD and BD models, as a function of the edge length (Figure~\ref{fig:shape_properties}b). The CD model, trained to treat aligned and closure as identical, produces linearly increasing closure as the edge length
increases. The BD model, trained to focus on background color, produces a flat function of closure with edge length.
Thus, when a model necessarily treats aligned and complete as identical, it produces
a monotonic ramp with edge length. When a model has no constraint on how it treats the different types
of images, it fails to produce closure at any edge length. We therefore use these
curves as signatures of closure and failure to exhibit closure, respectively.

Given that the CD model was trained to treat aligned images as
equivalent to complete triangles, regardless of edge length, it is surprising that 
internal representations of the model remain sensitive to edge length, as evidenced
by increasing closure with edge length.
Because the task requires determining how edges align in Gestalt shapes, 
it seems to be impossible for the CD model not to preserve information about edge 
length, albeit task irrelevant.  This feature of the model is consistent with
findings that human performance also varies as a continuous, monotonic function of the
edge length, whether the behavioral measure of closure is discrimination 
threshold  \citep{RingachShapley1996},  search latency \citep{elder1993effect},
or memory bias \citep{Holmes1968}.
Similarly, neurons in area 18 of the visual cortex of alert monkeys responding to illusory
contours show an increased strength of response as the edge length increases \citep{von1984illusory}.
These empirical results give further justification to treating the profile
of the CD model in Figure~\ref{fig:sanity} as a signature of closure.

\subsection*{The role of natural image statistics}

We now turn to the main focus of our modeling effort: to evaluate the hypothesis
that natural image statistics, in conjunction with a convolutional net architecture, 
are necessary and sufficient to obtain closure in a neural net.   We perform a series 
of simulations that provide converging evidence for this hypothesis.
Each experiment compares a \emph{base} model to a model that varies in a single aspect, 
either its architecture or training data.
Our base model is a state-of-the-art pretrained image classification network, 
Inception~\citep{Szegedy2016}. 

\begin{figure}[tb!]
    \begin{subfigure}{.24\textwidth}
  \centering
    \includegraphics[width=1.\linewidth]{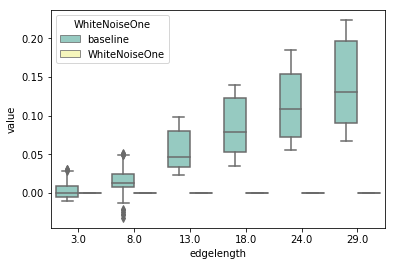}
  \caption{}
  \label{fig:7a}
\end{subfigure}
\begin{subfigure}{.24\textwidth}
  \centering
    \includegraphics[width=1\linewidth]{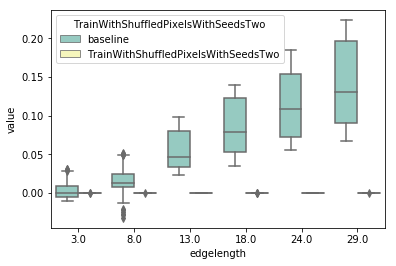}
  \caption{}
  \label{fig:7b}
\end{subfigure}
\begin{subfigure}{.24\textwidth}
  \centering
      \includegraphics[width=1\linewidth]{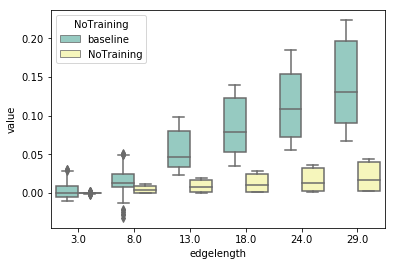}
  \caption{}
  \label{fig:7c}
\end{subfigure}
\begin{subfigure}{.24\textwidth}
  \centering
    \includegraphics[width=\linewidth]{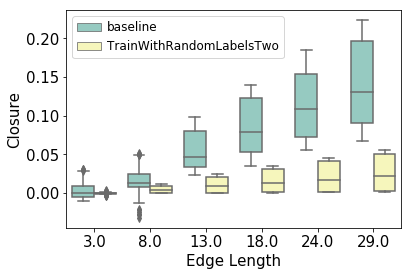}
  \caption{}
  \label{fig:7d}
\end{subfigure}
\begin{subfigure}{.24\textwidth}
  \centering
    \includegraphics[width=1\linewidth]{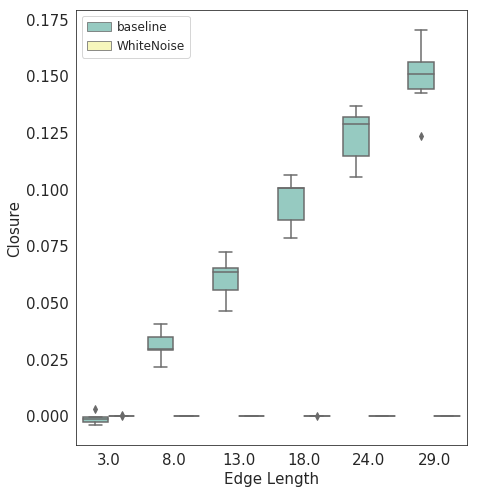}
  \caption{}
  \label{fig:7e}
\end{subfigure}
\begin{subfigure}{.24\textwidth}
  \centering
    \includegraphics[width=1\linewidth]{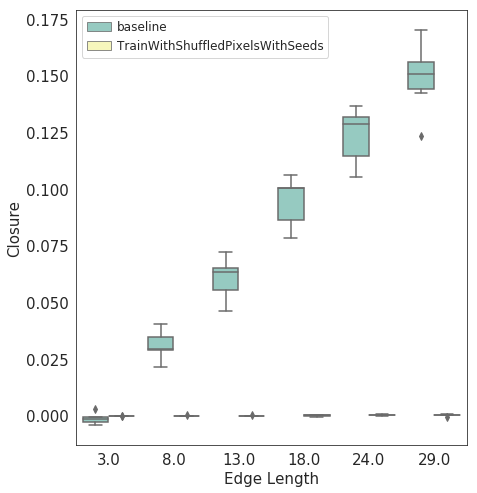}
  \caption{}
  \label{fig:7f}
\end{subfigure}
\begin{subfigure}{.24\textwidth}
  \centering
    \includegraphics[width=1\linewidth]{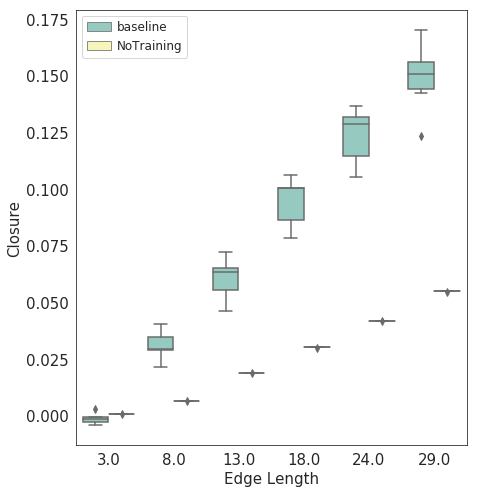}
  \caption{}
  \label{fig:7g}
\end{subfigure}
\begin{subfigure}{.24\textwidth}
  \centering
    \includegraphics[width=1\linewidth]{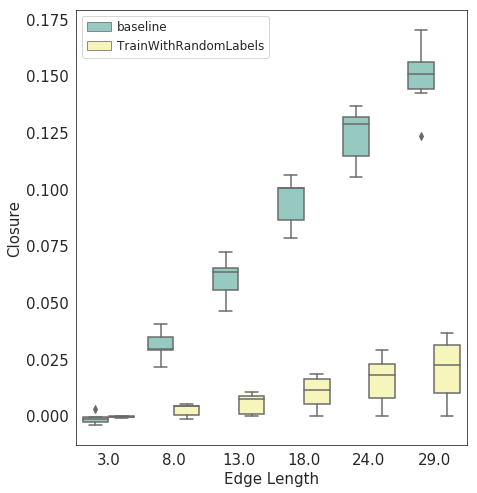}
  \caption{}
  \label{fig:7h}
\end{subfigure}
    \caption{
    Exploration of how closure is influenced by various aspects of the neural
    net. We test Inception with 1000-classes (panels a-d) and a smaller ConvNet
    architecture with 3, 6, or 9 classes (panels e-h).
    Each graph compares a standard ConvNet architecture trained on natural images to
    an alternative: 
    (a) comparing standard Inception to white-noise trained model, 
    (b) comparing standard Inception to model trained on shuffled pixels,
    (c) comparing standard Inception to untrained model,
    (d) comparing standard Inception to model trained on shuffled labels,
    (e) comparing small ConvNet to white-noise trained model,
    (f) comparing small ConvNet to model trained on shuffled pixels,
    (g) comparing small ConvNet to untrained model,
    (h) comparing small ConvNet to model trained on shuffled labels,
    }
    \label{fig:results}
\end{figure}

\subsubsection*{Natural images versus white noise}

Figure~\ref{fig:results}a compares the base model to an identical architecture trained 
on white noise images. The base model and white-noise net share not only the same architecture, but also training
procedure and initial weight distribution; they differ only in that the white-noise net does not benefit from natural 
image statistics. Nonetheless, the network has the capacity to learn a training
set of 
1.2 million  examples (same number as normal Imagenet training set) from
1001 randomly defined classes.
Each pixel in these images is sampled from a uniform $[-1,+1]$ distribution, the range of values that 
the model has for natural images after preprocessing.
The base model shows a closure effect: a monotonic increase in $\bar{\mathcal{C}}$ 
with edge length, whereas the white-noise net obtains $\bar{\mathcal{C}}=0$ regardless
of edge length. 
Performing a two-way analysis of variance with stimulus triple as the between-condition
random factor, we find a main 
effect of model ($F(1, 1188)=2507, p<0.0001$), 
a main effect of edge length ($F(5,1188)=254, p<0.0001$), 
and an interaction ($F(5,1188)=254, p<0.0001$).

\subsubsection*{Original images versus shuffled pixels}


Training on white noise might be considered a weak comparison point because the
low-order statistics (e.g., pairs of adjacent pixels) are quite different from those natural images.
Consequently, we tested an input variant that looks quite similar to white noise to
the human eye but matches pixelwise statistics of natural images:  images whose pixels
have been systematically shuffled
between image locations. While these shuffled images contain the same information
as natural images, the rearrangement of pixels not only prevents the human eye
from detecting structure but also blocks the network from learning structure
and regularities due to the local connectivity of the receptive fields.
Nonetheless, large overparameterized neural networks like Inception
have the capacity to learn the shuffled-pixel training set, although they will not generalize to new examples \citep{zhang2016}.

Figure~\ref{fig:results}b shows that Inception trained on shuffled pixels does not
obtain a closure effect. 
Performing a two-way analysis of variance, we find a main 
effect of model ($F(1,888)=1249.6, p < .0001$), 
a main effect of edge length ($F(5,888)=253.3, p < .0001$), 
and an interaction ($F(5,888)=126.7, p < .0001$).

\subsubsection*{Trained versus untrained networks}
Our white-noise and shuffled-pixel experiments indicate that training on
corrupted inputs prevents closure. Now we ask whether an untrained network
naturally exhibits closure, which is then suppressed by training in the case of 
corrupted images.

Figure~\ref{fig:results}c compares our base model to the same model prior to training,
with random initial weights. 
The untrained model exhibits a weaker closure effect
as indicated by an interaction between condition and edge length
($F(5,1188)=166.9, p < .0001$). Averaging over edge lengths, the magnitude of the 
random-weight closure effect is nonzero ($t(599)=19.7, p < 0.0001$), indicating that
some amount of closure is attributable simply to the initial architecture
and weights. 
This finding is not entirely surprising as researchers have
noted the strong inductive bias that spatially local connectivity and 
convolutional operators impose on a model, making them effective as feature
extractors with little or no data \citep{ulyanov2018deep,Zhang2020}. 
In the Supplementary Materials, we show that the amount of training required for the network
to reach its peak $\bar{\mathcal{C}}$ is fairly minimal, about 20 passes
through the training set, about $1/6$th  of the training required for the network
to reach its asymptotic classification performance. 



\subsubsection*{Systematic versus shuffled labels}

We have argued that the statistics of natural image data are necessary to obtain
robust closure, but we have thus far not explored what aspect of these statistics
are crucial. Natural image data consist of \{image, label\} pairs, where there
is both \emph{intrinsic} structure in the images themselves---the type of 
structure typically discovered by unsupervised learning algorithms---and \emph{associative} structure in 
the systematic mapping between images and labels. Associative structure is crucial 
in order for a network to generalize to new cases.

In Figure~\ref{fig:results}d, we compare our base model with a version trained
on shuffled labels, which removes the associative structure. Our model has the
capacity to memorize the randomly shuffled labels, but of course it does not generalize.
The shuffled-label model exhibits a weaker closure effect
as indicated by an interaction between condition and edge length
($F(5,1188)=143.0, p < .0001$). Averaging over edge lengths, the magnitude of the 
shuffled-label closure effect is nonzero ($t(599)=18.5, p < 0.0001$), indicating that
some amount of closure is attributable simply to intrinsic image structure.
We conjecture that the network must extract this structure in order to 
compress information from the original
$150\times150\times3$ pixel input into the more compact 2048-dimensional embedding,
which will both allow it
to memorize idiosyncratic class labels and---as a side effect---discover
regularities that support closure.
By this argument, supervised training on true labels
further boosts the network's ability to extract structure meaningfully related
to class identity. This structure allows the network to generalize to new
images as well as to further support closure.

We chose to eliminate associative structure by shuffling labels, but an alternative
approach might be to train an unsupervised architecture that uses only the input images,
e.g., an autoencoder. We opted not to explore this alternative because
label shuffling was a more direct and well-controlled manipulation; it allows us
to re-use the base model architecture as is.



\subsection*{Replication on simpler architecture}

To examine the robustness of the results we've presented thus far,  we conducted a 
series of simulations with a smaller, simpler architecture. This architecture has three output classes,
chosen randomly from the ImageNet data set, and three layers, each consisting of a convolutional mapping followed
by max pooling. We train 8-10 networks with the same architecture and different weight initializations.

Figures~\ref{fig:7e}-\ref{fig:7h} show closure results for the simple architecture that correspond
to the results from the larger architecture in Figures~\ref{fig:7a}-\ref{fig:7d}.
This simple architecture produces the same pattern of closure effects as the larger Inception model,
suggesting that closure is robust to architecture. Closure also appears to be robust to stimulus
image diversity: Inception is trained on images from 1000 distinct classes; the simple net is 
trained on images from only 3 classes. However, we have observed lower bounds on the required diversity:
When we train either model on one example per class, closure is not obtained. 

\subsection*{The role of convolutional operators and local connectivity}

\begin{figure}[b!]
    \centering
    \includegraphics[width=.45\linewidth]{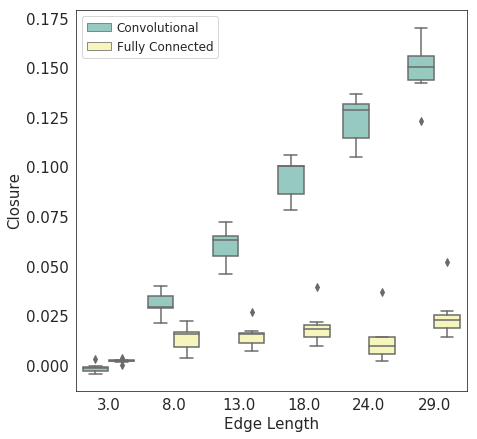}
    \caption{Exploring closure on convolutional versus fully connected architectures. Only the convolutional
    net achieves a closure effect, as indicated by the nonzero slope of the edge length vs.\ closure function. 
     }
    \label{fig:FCvsConv}
\end{figure}

Deep networks have become successful in vision tasks due to adopting some basic architectural
features of the mammalian visual system, specifically, the assumptions of local connectivity and equivariance
\citep{Fukushima1983}. Local connectivity in a topographic map indicates that a detector in one region of the 
visual field receives input only from its local neighborhood in the visual field.   Equivariance indicates that
when presented with the same local input, detectors in different parts of the topographic map respond similarly.
These properties are attained in deep nets via convolutional architectures with weight constraints.

To evaluate the role of these architectural constraints, we compare a ConvNet with the generic
alternative, a \emph{fully connected} architecture (\emph{FCNet}) with dense (non-local) connectivity and no built in
equivariance. Because FCNets do not perform as well on complex vision tasks, we were unable to
train an FCNet on the full ImageNet data set to achieve performance comparable to our 
baseline ConvNet. Without matching the two architectures on performance, any comparison confounds
architecture and performance. Consequently, we trained small instances of both architectures on just three
randomly selected classes from ImageNet, allowing us to match the 
ConvNet and FCNet on
performance. We replicated this simulation 7 times for robustness. 

Figure~\ref{fig:FCvsConv} compares the closure effect for ConvNets and FCNets.
The penultimate layer of representation is used to assess closure for both architectures.  
While the ConvNet evidences closure, the  FCNet does not. Taking this finding together
with the fact that the untrained ConvNet exhibits some degree of closure 
(Figures~\ref{fig:7c} and \ref{fig:7g}), we infer that some aspect of the ConvNet structure
facilitates the induction of closure.

\subsection*{Levels of representation and closure}

\begin{figure}[b!]
    \centering
    (a)
    \includegraphics[width=.45\linewidth]{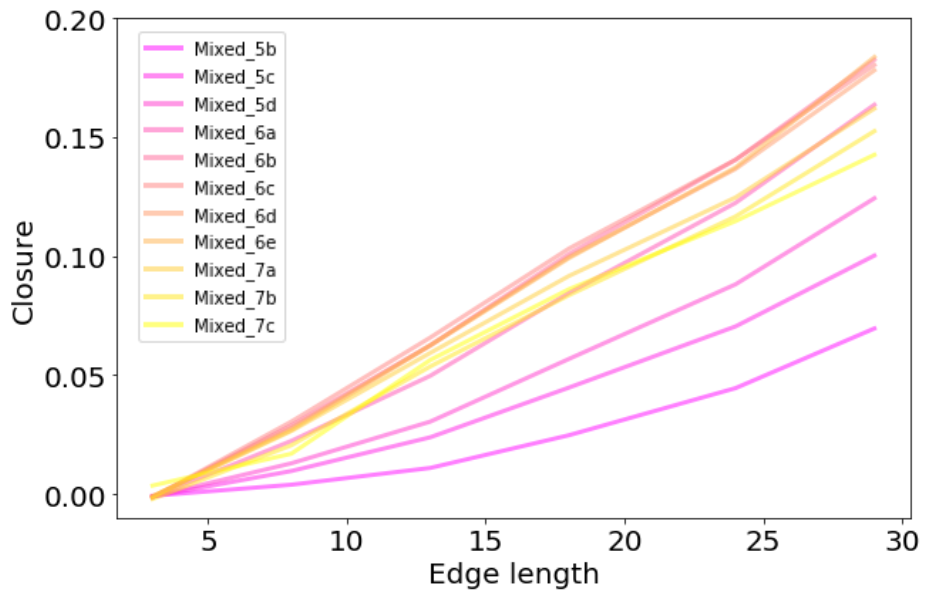}
    (b)
        \includegraphics[width=.45\linewidth]{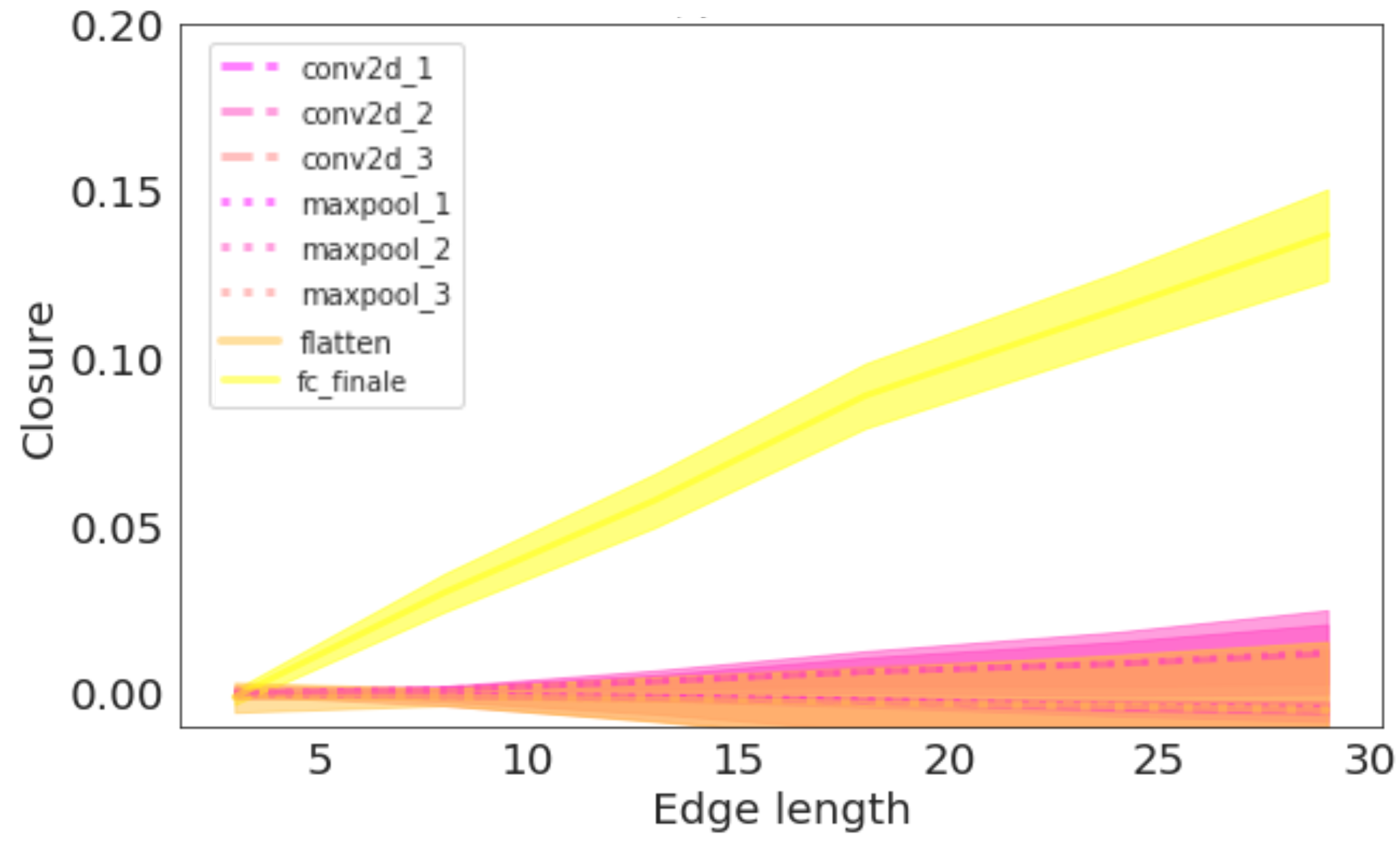}
    \caption{(a) The closure effect for the final eleven layers of the Inception architecture. Previously,
    we assessed closure only at layer `Mixed\_7c', but the lower layers also show varying degrees of closure.  
    (b) The closure effect for each layer of the small ConvNet. Previous results have
    read out from the `fc\_finale' layer. In both graphs, variability over images in the strength of closure
    is shown with uncertainty shading.
    }
    \label{fig:AllLayers}
\end{figure}

Thus far, we have investigated closure at the penultimate layer of a network, on the assumption
that this representation would be the most abstract and therefore most likely to encode object shape. However the deep Inception architecture we tested
has 16 major layers, some of which involve multiple convolutional transformation and pooling operations.
A priori, observing closure in early layers seems unlikely because the receptive fields of neurons in 
these layers have spatially constrained receptive fields, and closure requires the registration of 
Gestalt properties of the shapes. (Our test of closure will not false trigger based on local visual
edge similarity because we compare images with distinct $\theta_{\text{global}}$.)

In Figure~\ref{fig:AllLayers}a, we show the closure effect for representations in the last eleven layers of
Inception.  `Mixed\_7c' is the layer whose representation we have previously reported on.
The graph legend is ordered top to bottom from shallowest to deepest layer.
While all of the eleven layers show closure, closure is weaker for the shallower layers, labeled `Mixed\_5',  
than the deeper layers, labeled `Mixed\_6' and `Mixed\_7'. We do not have a definitive explanation for why
the effect is slightly weaker in the  deeper `Mixed\_7' layers than in the shallower `Mixed\_6' layers, though 
we suspect it is a consequence of training the model for classification.
Classification does not require any information other
than class labels to be transmitted through the network. Consequently, the net is not encouraged to preserve 
a shape representation through all layers, and the net possibly discards irrelevant shape information in order
to optimize inter-class discrimination.

In Figure~\ref{fig:AllLayers}b, we depict the closure curves for layers of the simple net, from the shallowest
hidden layer,
`conv2d\_1', to the penultimate layer, `fc\_finale'.   For this architecture, only the penultimate layer 
shows closure. In the penultimate layer, each neuron can respond to information anywhere in the visual field.

Consistent across Inception and the simple net, representations at the shallow layers are not
sufficiently abstract to encode Gestalt closure.
This follows from the local feedforward connectivity of the architectures and gradual collapsing (pooling) 
of information across increasingly wider receptive fields at deeper stages of both architectures.

Although filling-in phenomena are observed in early stages of visual cortex \citep{von1984illusory}, it's possible
that these effects are not causally related to Gestalt perception or are due to feedback projections, which our
models lack. But our results are otherwise consistent with the view that lower stages of neural nets capture
spatially local, low-order statistics whereas higher stages capture spatially global, high-order statistics
\citep{netdissect2017,mozer1991}.

\section*{Discussion}

Our work follows a long tradition in computational modeling of using neural network models to
explain qualitative aspects of human perception
\citep[e.g.,][]{rumelhart1988learning,mozer1991}.
The strength of computational methods
over behavioral or neuroscientific methods as an investigative tool is that models can be
precisely manipulated to determine the specific model properties and inputs that are necessary and
sufficient to explain a phenomenon.  Further, we can do more than merely observe a model's input-output
behavior; we can probe its internal representations and directly determine what it is computing.

We began with the conjecture that Gestalt laws need not be considered as primitive assumptions
underlying perception, but rather, that the laws themselves may arise from a more fundamental
principle: adaptation to statistics of the environment. We sought support for this conjecture
through the detailed study of a key Gestalt phenomenon, closure.  
Using a state-of-the-art deep neural network model that was pretrained to
classify images, we showed that in the model:
\begin{itemize}
\item \emph{Closure depends natural image statistics.} Closure is obtained
for large neural networks trained to classify objects, and even for a smaller net
trained to discriminate only a few object classes, but it is not obtained when a net is
trained on white  noise images or shuffled-pixel images. While shuffled-pixels have the same 
statistics as  natural images, networks with local receptive fields are unable to extract 
spatially local structure due to the fact that the pixel neighborhood
has been randomly dispersed in a shuffled image.
\item \emph{Closure depends on the architecture of convolutional networks}.
The extraction of image regularities is facilitated by two properties of ConvNets:
spatially local receptive fields and equivariance. Fully connected
networks, which lack these forms of inductive bias, do not obtain closure. The inductive
biases are sufficiently strong that even an untrained ConvNet obtains a
weak degree of closure.
\item \emph{Closure depends on learning to categorize the world in a meaningful way.}
Networks trained to associate images with their correct categorical label produce much
larger closure effects than networks trained to associate images with random labels.
In the former case, the labels offer a clue about what features should be
learned to systematically discriminate categories \citep{Lupyan2012}.
In the latter case, the labels
misdirect the net to discover features that, by chance, happen to
be present in a random collection of images that were assigned the same label.
\end{itemize}

Our simulation experiments suggest that these dependencies are both necessary and sufficient
for a computational system to produce closure. The system need not have innate sensitivity to closure,
nor does it need to \emph{learn} closure per se. Rather, closure emerges as a byproduct of learning 
to represent real-world objects and categories.

Although we made this argument specifically for closure, the same argument applies to other
Gestalt laws that have underpinnings in natural scene statistics
\citep{Brunswik1953,ElderGoldberg2002,Geisler2001,Kruger1998,Sigman2001}.
Given structure in the environment and the existence of powerful learning architectures and
mechanisms, one need not consider the Gestalt laws as primitives. Rather, the Gestalt laws can emerge via
adaptation to the statistical structure of the environment.

One limitation of our work is that we do not claim our trained neural net classifier is a cognitive
model, i.e., a model of the cognitive processes involved in biological perception.  At some very coarse
level of analysis---e.g., the fact that the model learns from experience to classify naturalistic images---the
model does replicate human abilities. However, we have not tried to explain detailed and specific experimental
phenomena in the way one typically does in cognitive modeling. 
A phenomenon we most likely cannot explain in our model is human performance in the
classification image paradigm---the paradigm discussed earlier in which noisy stimuli are averaged together
to infer illusory contours. A pretrained neural net much like the one we studied does 
not show the same effect as human subjects \citep{Baker2018}. However, the model is not inconsistent with
the phenomenon; rather it would have to be expanded to explain a full range of phenomena.
In this case, the expansion might involve a different input representation,
such as edge detectors, or feedback projections to achieve better biological realism
\citep{Spoerer2017}.

We have focused our study on modal completion, which involves the detection of illusory shapes. 
Amodal completion, which involves the detection of occluded shapes, provides a further opportunity to
evaluate the consequences of learning natural scene statistics.
There is some debate whether modal and amodal completion rely on the same visual mechanisms in humans
\citep{Anderson2007,Kellman1998}.
In our model, modal completion likely arises because explicit perceptual evidence about
the  edges of an object in a natural scene is occasionally absent; thus, natural image perception demands inferring 
the missing features. In our model, amodal completion would arise for the analogous reason:  the presence
of partially occluded objects where missing edge features again must be inferred.  The case for occlusion in
natural scenes is much stronger than the case for missing features, and thus an adaptation-based account of
amodal completion seems quite natural. Indeed, the \citet{Zemel2002} study (Figure~\ref{fig:zemel}) provides
clear evidence that completion can be learned under occlusion as a consequence of experience. 

In the late 1980s, the Connectionist movement focused on the role of learning in perception and cognition, 
demonstrating that abilities that one might previously have thought to have been built into a cognitive system 
could emerge as a consequence of simple learning rules. Most connectionist architectures were fairly 
generic---typically fully connected neural networks with one hidden layer. While such architectures showed promise,
it was far from clear that these architectures could scale up to human-level cognitive abilities. 
Modern deep learning architectures have clearly scaled in a manner most did not imagine in the 1980s.
Large language models show subtle linguistic skills and can express surprisingly broad knowledge about the 
world \citep{raffel2019}; large vision models arguably match or surpass human abilities to label images \citep{xie2019}.
Our work suggests that in the modern era, the combination of a sophisticated neural architecture (e.g., a ConvNet)
and scaling the size of models may be sufficient to broaden the range of cognitive phenomena that are emergent from
more basic principles of learning and adaptation. Our work bridges the gap between analyses indicating that perceptual
mechanisms are consistent with natural scene statistics \citep{Burge2010,Brunswik1953} and claims that statistical
learning is essential to understanding human information processing \citep{Frost2019}.
The synthesis leads to a view of the human perceptual system that
is even more elegant than the Gestaltists imagined: a single principle---adaptation to  the statistical structure of 
the environment---might suffice as fundamental.



\section*{Methods}


\subsection*{Models}

In our large simulation experiments, we leverage a state-of-the-art, 
pretrained image classification network, \emph{Inception}, trained on the 1000-class ImageNet data
set \citep{Szegedy2016}. Input to this model is a $150\times150$ pixel color (RGB) image, and output 
a 1001-dimensional activation vector whose elements indicate the probability that
the image contains the corresponding object class.
(The ImageNet task involves 1000 classes; the additional class is a `none of the above' category.)
Inception has been trained on 1.2 million images from the ImageNet data set \citep{imagenet2009}
to assign each image to one of a thousand object classes. We train with standard data augmentation
methods including: horizontal flips, feature-wise normalization, aspect ratio adjustment, shifts, and color distortion. 
In most simulations, we read out representations from the penultimate layer of the net, known as Mixed\_7c,
which consists of a 2048-dimensional flat vector.
The penultimate layer of a deep net is commonly assumed to embody
a semantic representation of the domain (visual objects). 
For example, in transfer learning and few-shot learning models, this layer is
used for encoding novel object classes \citep[e.g.,][]{Scott2018,Yosinski2014}.
One of our experiments reads out from earlier layers of the network.


We also explore a \emph{simple convolutional} architecture consisting of three pairs of alternating convolutional and 
max pooling  layers followed by a fully connected layer and a single output unit trained to discriminate between three
classes, randomly chosen from ImageNet.  A \emph{fully connected} variant of the architecture replaces the convolutional
and pooling blocks with fully connected layers.  For these simple models, the embedding is the penultimate layer
of the network. 

For the sanity-check experiments (CD and BD models), we used the simple convolutional architecture with the three 
output classes with a single output which  performs a  binary discrimination (disordered versus complete and 
aligned for CD; black backgrounds versus white backgrounds for BD).  The CD and BD models are trained on 75\% 
of the 768 distinct distinct complete-aligned-disordered triples; the remainder form 
a validation set, which reaches 100\% accuracy and is used for evaluating the model. 
Five replications of the CD and BD models are trained with different random
initial seeds to ensure reliability of results.

Further details on all models are provided in the Supplementary Materials.

\subsection*{Stimuli}


We compare three critical conditions (Figure~\ref{fig:kanizsa}c-e): complete triangles, 
triangle fragments with aligned corners, and fragments with disordered corners. 
Each stimulus is rendered in a $150\times150$ pixel image and the Euclidean
distance between vertices is 116 pixels.
Rather than testing models with more elaborate images 
(e.g., Figures~\ref{fig:kanizsa}a,b), 
we chose to use the simplest images possible that could evoke closure 
effects, for two reasons. First, with more complex 
and naturalistic images, we found that it was difficult to control for 
various potential confounds.
Second, complex examples like the  Kanizsa triangle potentially involve both
modal completion (the white triangle in the foreground of Figure~\ref{fig:kanizsa}a) and amodal completion (the outline triangle 
and the solid  black disks behind the  foreground triangle) completion). 
We wished to simplify and focus specifically on
the Gestalt principle of closure.

We manipulated various properties of the stimuli, as depicted in 
Figure~\ref{fig:shape_properties}. For all conditions, 
the stimuli varied in the global orientation of the triangle or fragments, 
which we refer to as $\theta_{\mathrm{global}}$,
the \emph{background} (light on dark versus dark on light), 
and the \emph{position} of the object center in the image. 
For the disordered condition, we varied the orientation of the corners 
with respect the orientation of corners in the aligned-fragment condition, 
which we refer to as $\theta_{\mathrm{local}}$.
And finally,  for the disordered and aligned conditions, we varied the length 
of the edges extending from the fragment corners, which we refer to 
as \emph{edge length}.

Edge length is centrally related to the phenomenon of interest. Edge  
length, or equivalently, the gap between corners, influences
the perception of closure, with smaller gaps leading to stronger closure
\citep{elder1993effect,JAKEL20163}. The remaining properties---background
color, local and global orientation, and image position---are manipulated to
demonstrate invariance to these properties. If sensitivity to any of 
these properties is observed, one would be suspicious of the generality of
results. Further, these properties must be varied in order to avoid a
critical confound: the  complete image (Figure~\ref{fig:kanizsa}d)
shares more pixel overlap with the aligned fragments (Figure~\ref{fig:kanizsa}c)
than with the disorderd fragments (Figure~\ref{fig:kanizsa}d). We therefore
must ensure that any similarity of response between complete
and aligned images is not due to pixel overlap. We accomplished this
aim by always comparing the response to complete and fragment images 
that have different $\theta_\mathrm{global}$ and different image 
positions. However, when comparing representations, we always match
the images in background color because neural nets tend to show
larger magnitude responses to brighter images. 

The background color has two levels, black and white. 
The position is varied such that the stimuli could be centered on the middle 
of the image or offset by $-8$ pixels from the center in both $x$ and 
$y$ directions, resulting in two distinct object locations.
The global orientation is chosen from eight equally
spaced values from $0^\circ$ to $105^\circ$. (Symmetries make additional
angles unnecessary. A $120^\circ$ triangle is identical to a 
$0^\circ$ triangle.) 
The local orientation of the disordered corners is rotated from the
aligned orientation by $72^\circ$, $144^\circ$, $216^\circ$, or $288^\circ$.
The edge length is characterized by the length of an edge emanating from 
the vertex; we explored six lengths: 3, 8, 13, 18, 24, and 29 pixels, which
corresponds to removal of between 95\% and 50\% of the side of a complete triangle 
to form an aligned image.
These manipulations result in $2\times2\times8 = 32$ complete triangles,
$2\times2\times8\times6=192$ aligned fragments, and
$2\times2\times8\times6\times4=768$ disordered fragments, totalling
992 distinct stimulus images.

\subsection*{Quantitative measure of closure}

\citet{elder1993effect} studied closure via a visual-search task that 
required human participants to discriminate among simple shapes like
those in Figure~\ref{fig:shape_properties}.  Although we could, in principle,
ask a network to perform this task, additional assumptions would be required
about response formation and initiation. Instead, we claim simply that
the difficulty of the task depends on the similarity of internal
representations: if \{complete, aligned\} pairs are more similar to one another
than \{complete, disordered\} pairs, discrimination ability and response latency should
be longer by any sensible biologically-plausible read out process
\citep[e.g.,][]{RatcliffMcKoon2008}.
This claim allows us to focus entirely on the representations
themselves, via a quantitative measure of closure:
\[
\mathcal{C}_i = s(f(\boldsymbol{a}_i), f(\boldsymbol{c}_i)) -
s(f(\boldsymbol{d}_i), f(\boldsymbol{c}_i)) , 
\]
where $i$ is an index over matched image triples consisting of a complete
triangle ($\boldsymbol{c}_i$),  aligned fragments ($\boldsymbol{a}_i$),  and 
disordered fragments ($\boldsymbol{d}_i$); $f(.) \in \mathbb{R}^m$ is the 
neural net mapping from an input image in $\mathbb{R}^{150\times150}$ to an
$m$-dimensional embedding, and $s(.,.)$ is a similarity function
(Figure~\ref{fig:shape_properties}). Consistent
with activation dynamics in networks, we use a standard similarity measure, the cosine
of the angle between the two vectors,\footnote{We assume 
$s(\boldsymbol{x}, \boldsymbol{y}) = 0$ if both $|f(\boldsymbol{x})|=0$ and $|f(\boldsymbol{y})|=0$.
}
\[
s(\boldsymbol{x}, \boldsymbol{y}) = 
\frac{f(\boldsymbol{x}) f(\boldsymbol{y})^{\mathrm{T}}   }{|f(\boldsymbol{x})|~ |f(\boldsymbol{y})|}\;.
\]
The triples are selected such that $\boldsymbol{a}_i$ and $\boldsymbol{d}_i$
are matched in $\theta_{\mathrm{global}}$ position, both differ 
from $\boldsymbol{c}_i$ in $\theta_{\mathrm{global}}$,
and all three images have the same background color (black or white).
These constraints ensure that there is no more pixel overlap (i.e.,
Euclidean distance in image space) between
complete and aligned images than between complete and disordered images.

We test 768 triples by systematically pairing each
of the 768 distinct disordered images with randomly selected aligned and
complete images, subject to the constraints in the previous paragraph. 
Each of the 192 aligned images in the data set is repeated
four times, and each of the 32 complete images is repeated 24 times.

We compute the mean closure across triples, $\bar{\mathcal{C}} \in [-1,+1]$. 
This measure is $+1$ when the complete image yields a representation identical 
to that of the aligned image and orthogonal to that of the disordered image. 
These conditions are an unambiguous indication of closure because the model
cannot distinguish the complete triangle from the aligned fragments.
Mean closure $\bar{\mathcal{C}}$ is $0$ if the
complete image is no more similar to the aligned than disordered images,
contrary to what one would expect by the operation of Gestalt grouping
processes that operate based on the alignment of fragments to construct a
coherent percept similar to that of the complete triangle.
Mean closure $\bar{\mathcal{C}}$ may in principle be negative, but we do not 
observe these values in practice.

Although our measure of representational similarity is common in the 
deep learning literature, the neuroimaging literature has suggested other
notions of similarity, e.g.,  canonical correlation analysis
\citep{hardle2007applied} and representational similarity analysis~\citep{kriegeskorte2008representational}. However, these measures
are primarily designed to compare signals of different dimensionality (e.g.,
brain-activity measurement and behavioral measurement).

\section*{Acknowledgements}
We are grateful to Mary Peterson for insightful feedback on an earlier draft of the paper, and to
Bill Freeman and Ruth Rosenholtz for helpful discussions about the research.
Special thanks to Corbin Cunningham for his advice on our experiment designs.

\bibliography{arxiv}
\bibliographystyle{apalike}


%

%

\newpage
\section*{Supplementary Materials}
\subsection*{Experimental setup details}

\paragraph{Simple network}
Each network has $n_c \in \{3, 6, 9\}$ classes (randomly chosen from Imagenet dataset) with $n_l \in \{3, 5, 7\}$ 
layers. These layers are either convolutional or fully connected. For convolutional networks (ConvNets), we iterate between convolutional and max-pooling layers $n_l$ times to predict $n_c$ classes. In each layer, the number of neurons increases by 16, except the final layer always has 512 neurons regardless of the $n_c$. For fully connected networks (FCNets), we first flatten the input image, then add $n_l$ fully-connected layers. All networks are learned using the RMSProp method~
with $0.001$ (for convolutional) $0.0001$ (for FCNet, smaller rate was critical for learning) learning rate for 100 epochs. 
The training dataset was prepared with standard data augmentation: feature-wise normalization, linear translation (0.02 range) and horizontal flips.

\paragraph{Sanity check network}
This network has the identical setup to the simple network above with 3 convolution layers ($nl = 3$) and 2 classes ($nc = 2$).

\paragraph{More complex network (Inception)}
This network uses more complex and more widely used
 InceptionV3 network architecture~\citep{Szegedy2016}. 
 This network was trained on 1.2 million ImageNet images, with similar augmentation to the simple network: horizontal flips, featurewise normalization, aspect ratio adjustment, shifts, and color distortion. It was trained to top-5 accuracy of 92\% over 120 epochs with a batch size of 4096. The learning rate and weights decay followed those of \cite{DBLP:journals/corr/abs-1805-08974}.
 A depiction of the architecture is shown on \url{https://microscope.openai.com/models/inceptionv3_slim}.

\begin{figure}[h!]
    \centering
\includegraphics[width=.7\linewidth]{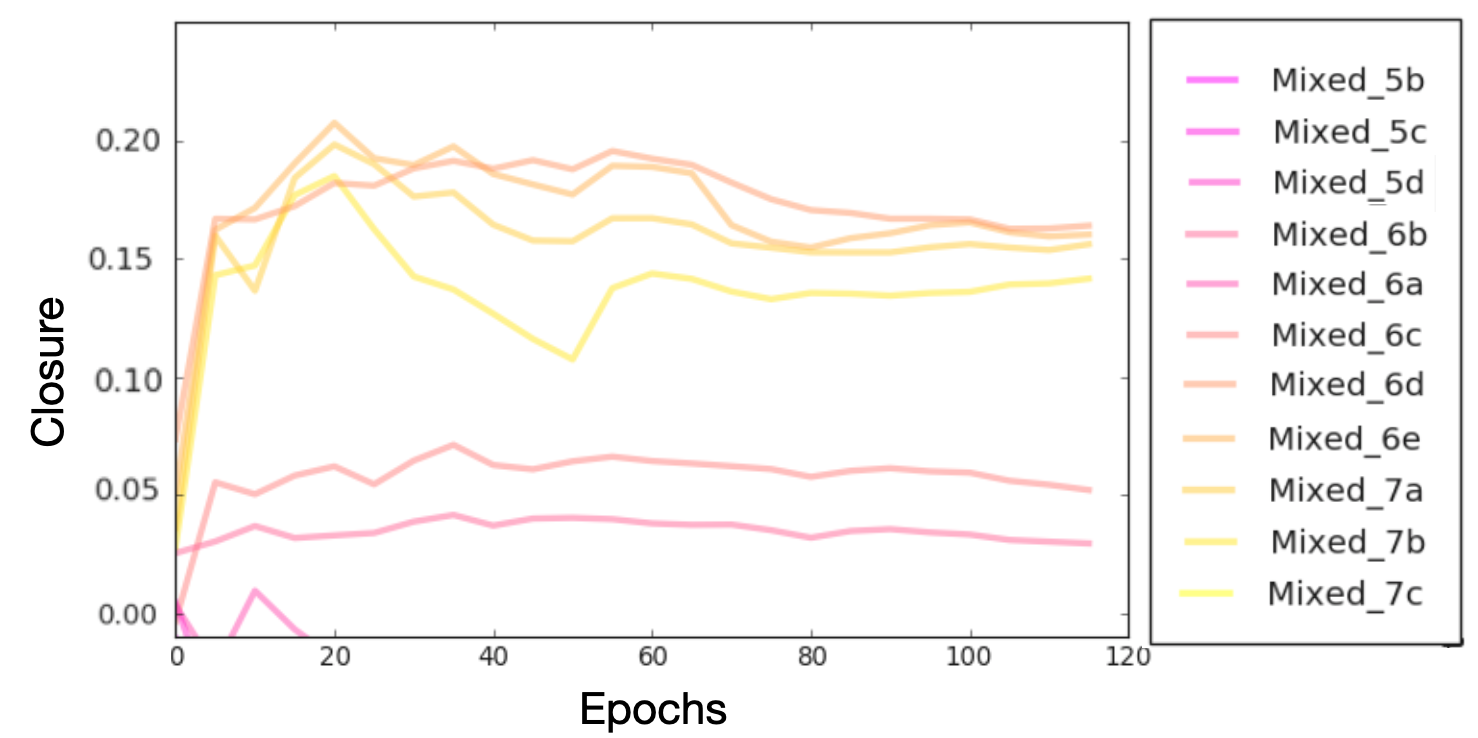}
    \caption{The closure effect in the last layer reaches its peak earlier in the training process, then decreases somewhat as it converges. Other layers seem to continue to fluctuate before converging. For degenerated networks, the effect is quickly forgotten. Showing results from Inception.}
    \label{fig:QlQconvQocc}
\end{figure}

\subsection*{The closure effect during training before convergence} 
We hypothesized that the closure effect would increase during training and will converge, similar to a typical validation accuracy curve.
This varies depending on the layer.
The closure effect reaches its peak earlier in the iteration, then fluctuates as it learns, then forgets the effect slightly as it converges in both simple network and Inception (Fig.~\ref{fig:QlQconvQocc}). This is typically observed in higher layers (e.g., Mixed 7a and above in Inception). In lower layers (e.g., Mixed 6d and lower), the closure effect increases then converges. 

On the other hand, networks trained with degenerate training data (e.g., shuffled pixels, shuffled labels) start with some closure effect, since untrained network exhibits some closure effect. However the closure effect drops immediately and stays close to zero in the duration of training. 

The rapid decrease of the closure effect in networks other than the baseline network aligns with our findings; in the process of trying to fit to random data, the network loses some of the initial feature extraction properties that it had had due to convolutional operations~\citep{ulyanov2018deep}, and the closure effect is also lost.

The fluctuation during learning before convergence is an interesting symptom, and may benefit from further study. This hints that the closure effect may reflect a prediction-related signal that can be useful (e.g., to determine stopping conditions).

\begin{figure}
    \centering
    \includegraphics[width=1.\linewidth]{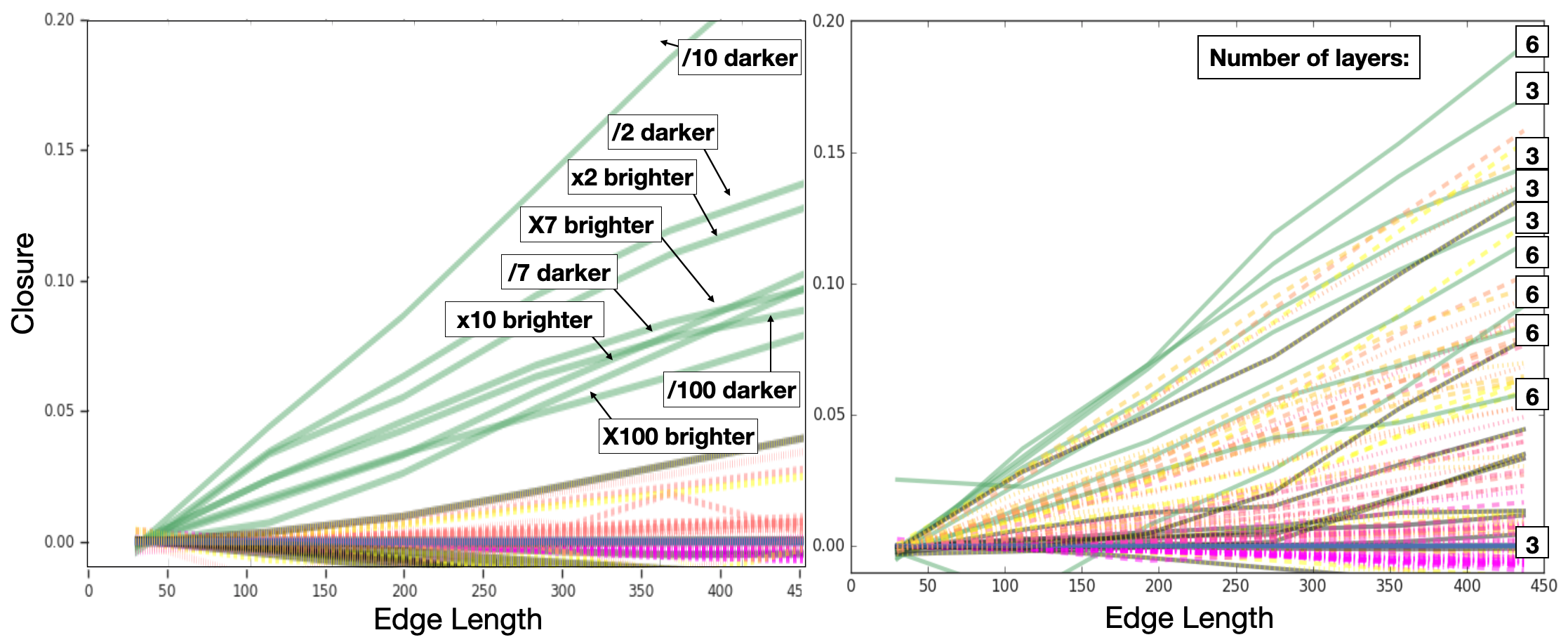}

    \caption{Varying brightness of stimuli and the number of layers: The closure effect does not have clear pattern of change as images become brighter/darker or as the depth of network and/or number of classes trained change.}
    \label{fig:numlayer}
\end{figure}

\subsection*{The closure effect is uninfluenced simple input manipulations} 
In this experiment, we try to invalidate a hypothesis that a seemingly meaningless input manipulation on the training data (e.g., image brightness) will arbitrarily influence the closure effect.
For example, we should not be able to increase/decrease the closure effect by simply making training images brighter/darker (i.e., multiplying/dividing images with constant).
%

There is no strong pattern between closure effects and brightness of the training images (Fig.~\ref{fig:numlayer}).
Note that the variance in each run reflects the amount of information lost by multiplying or dividing each pixel and saturating them (e.g., multiplying images with $10$ cause some images to be no longer identifiable). Naturally in conditions with no strong closure effect (e.g., shuffled pixels), there was no difference of the effect from brightness variation.

\begin{figure}
    \centering
    \includegraphics[width=1.\linewidth]{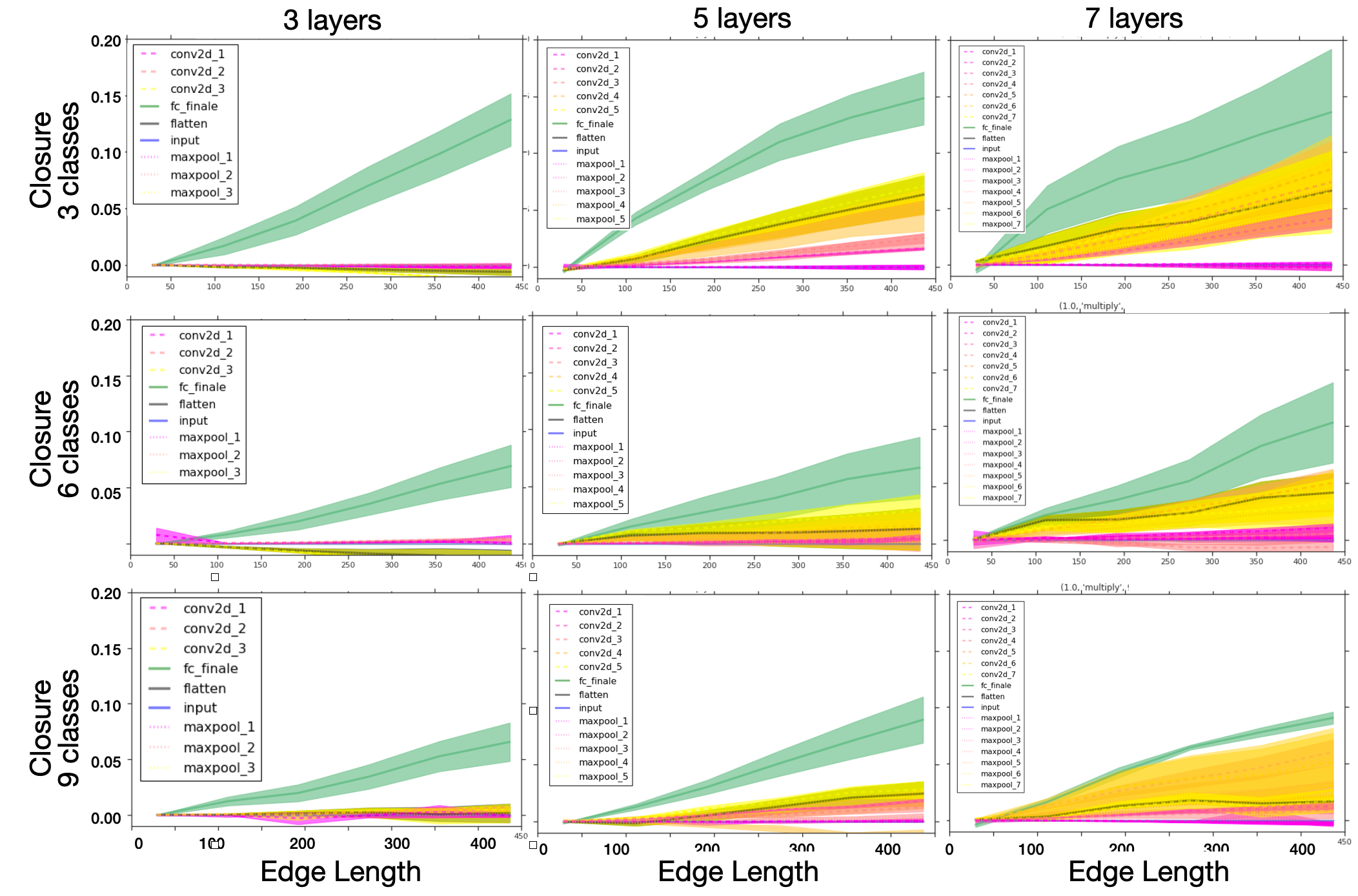}
    \caption{The closure effect for while changing number of classes, $nc$, and number of layers, $nl$. The effect does not seem to have clear pattern with respect to these factors.\label{fig:ncnl}}
\end{figure}

\subsection*{The closure effect as number of classes and number of layers vary.}

We discovered that there seems to be no strong correlation between the depth of the network and the closure effect (Figures~\ref{fig:ncnl}). 
Given that the optimal choice of the depth of a network is an unsolved problem, despite much work~\cite{lin2017does, ba2014deep}, and it is not always true that adding more layers improve the model's performance~\cite{ba2014deep}, it is reasonable that we cannot influence the closure effect by simply adding more layers.



\end{document}